\documentclass{article}

\usepackage[preprint]{neurips_2019}

\usepackage[utf8]{inputenc} 
\usepackage[T1]{fontenc}    
\usepackage{hyperref}       
\usepackage{url}            
\usepackage{booktabs}       
\usepackage{amsfonts}       
\usepackage{nicefrac}       
\usepackage{microtype}      
\usepackage[dvipsnames]{xcolor}
\usepackage{adjustbox}
\usepackage{tikz}
\usetikzlibrary{calc,positioning,patterns,shapes}
\makeatletter
\def\thanks#1{\protected@xdef\@thanks{\@thanks
        \protect\footnotetext{#1}}}
\makeatother

\title{The TCGA Meta-Dataset Clinical Benchmark}

\author{%
  Mandana Samiei\textnormal{\textsuperscript{1}}\quad
  Tobias W\"{u}rfl\textnormal{\textsuperscript{2}}\quad
  Tristan Deleu\textnormal{\textsuperscript{3}}\quad
  Martin Weiss \textnormal{\textsuperscript{3}}\\
  \textbf{Francis Dutil\textnormal{\textsuperscript{5}}\quad
  Thomas Fevens\textnormal{\textsuperscript{1}}\quad
  Geneviève Boucher\textnormal{\textsuperscript{3,\,4}}\quad
  Sebastien Lemieux\textnormal{\textsuperscript{3,\,4}}}\\
  \textbf{Joseph Paul Cohen\textnormal{\textsuperscript{3}}}\quad
  \thanks{
  \textsuperscript{1}\,Concordia University, %
  \textsuperscript{2}\,Friedrich-Alexander-Universit\"{a}t Erlangen-N\"{u}rnberg, %
  \textsuperscript{3}\,Universit\'{e} de Montr\'{e}al, %
  \textsuperscript{4}\,Institute for Research in Immunology and Cancer. %
  \textsuperscript{5}\,Imagia. %
  Correspondance: \texttt{samiemandana@gmail.com}.}\\[1em]
  Mila -- Montreal, Canada\\
}

\begin{document}

\maketitle

\begin{abstract}
Machine learning is bringing a paradigm shift to healthcare by changing the process of disease diagnosis and prognosis in clinics and hospitals. This development equips doctors and medical staff with  tools to evaluate their hypotheses and hence make more precise decisions. Although most current research in the literature seeks to develop techniques and methods for predicting one particular clinical outcome, this approach is far from the reality of clinical decision making in which you have to consider several factors simultaneously. In addition, it is difficult to follow the recent progress concretely as there is a lack of consistency in benchmark datasets and task definitions in the field of Genomics. To address the aforementioned issues, we provide a clinical Meta-Dataset derived from the publicly available data hub called The Cancer Genome Atlas Program (TCGA) that contains 174 tasks. 
We believe those tasks could be good proxy tasks to develop methods which can work on a few samples of gene expression data. Also, learning to predict multiple clinical variables using gene-expression data is an important task due to the variety of phenotypes in clinical problems and lack of samples for some of the rare variables. The defined tasks cover a wide range of clinical problems including predicting tumor tissue site, white cell count, histological type, family history of cancer, gender, and many others which we explain later in the paper. Each task represents an independent dataset. We use regression and neural network baselines for all the tasks using only 150 samples and compare their performance. 
\end{abstract}

\section{Introduction}
\label{sec:introduction}

Most researchers develop new methods for one particular clinical task at a time, as an example, we can name survival prediction \citep{Wieringen:2019:SP}  or tumor cell type classification \citep{Lyu:2018:DLB}. This approach is far from the reality of clinical decision making in which you have to consider several clinical variables simultaneously which are often performed by doctors and clinical staff. In some cases, those prediction tasks are correlated with each other. For instance, 
by knowing about gender, alcohol history documents, and tumor tissue site, we can achieve a more reliable result of cancer mortality. To tackle a precise outcome for many clinical tasks, it's more practical to consider inter-related tasks. In addition, there are some tasks that do not have enough number of samples. By considering a few-shot learning regime, we leverage the samples from a collection of tasks and construct a prior knowledge for a general prediction.

Additionally, while there has been an extreme growth in Machine Learning research for Genomics, several barriers have slowed the progress. One particular barrier is the absence of global publicly available benchmark datasets in the field of gene expression works to evaluate the performance of models, facilitate reproducibility and allow the community to focus on a set of challenges. There are consistent public benchmarks in the field of computer vision for image classification tasks, for example, ImageNet \citep{ILSVRC15} and Large Scale Visual Recognition Challenge (ILSVRC) \citep{ILSVRC15} which allowed a significant improvement in the classification error from 25\% in 2011 to 95\% in 2017 for benchmark tasks. In contrast, practical progress in genomics applications has been difficult to measure due to variability and inconsistency in data sets and task definitions. Although there is a fairly well-known challenge called DREAM that can be found at \texttt{\href{http://dreamchallenges.org/}{http://dreamchallenges.org/}}, it has not imposed itself as a standard challenge such as ImageNet.

In this paper, we propose a collection of 174 TCGA benchmark clinical tasks that can be used in a multi-task learning framework. All the tasks are classification problems and the input space is gene-expression data with 20530 genes. We apply independent regression and neural models that will predict across any general clinical tasks. The purpose of each dataset that can be inferred by its name is predicting a clinical variable of a particular cancer type. For instance, one of the datasets is named 'gender-LAML' means we predict 'gender' for 'Acute Myeloid Leukemia' samples. 
 
We evaluate the performance of each clinical task using logistic regression and a neural network that will predict across corresponding cancer study. This framework is an initiative to develop Meta-Learning techniques that make use of the interrelated clinical tasks of gene-expression data to learn effective classification models despite the small sample size of each individual task.
The code is accessible at: 
\texttt{\href{https://github.com/mandanasmi/TCGA_Benchmark}https://github.com/mandanasmi/TCGA$\_$Benchmark}.

\subsection{Summary of Contributions}

This paper proposes a public Meta-Dataset that provides 174 defined clinical tasks. TCGA Meta-Dataset also includes a meta-dataloader which is available on the Github repository. The organization of the paper is as follows: Section~\ref{sec:related-works} presents related works to this paper, and Section~\ref{sec:task-definition} gives an overview of the public TCGA dataset and the defined tasks. Section~\ref{sec:data-loaders-for-TCGA-Meta-Dataset} explains the data-loaders available in the  TCGA Meta-Dataset. Section~\ref{sec:experimental-results} gives a detail of the experimental and training setup, the baselines we used, the evaluation method, the architecture as well as a comparison of models' performance on different tasks. In the end, Section~\ref{sec:conclusion-and-future-works} concludes the work and explains how we plan to extend this in future work. 

\vspace*{-0.5em}
\section{Related Works}
\label{sec:related-works}

A recent paper in Meta-Learning literature which is closely related to this work is Meta-Dataset by  \citep{Eleni:metadataset:2019} that offers an environment for training
and evaluating meta-learners for few-shot image classification by
evaluating various baselines and meta-learners on this Meta-Dataset.
This benchmark includes 10 datasets from various image classification tasks including ILSVRC-2012 (ImageNet) \citep{ILSVRC15}, Omniglot \citep{Lake:omniglot:2019}, Aircraft \citep{DBLP:journals/corr/MajiRKBV13}, MSCOCO \citep{10.1007/978-3-319-10602-1-48}, CUB-200-2011 (Birds) \citep{WahCUB2002011}, Describable Textures \citep{Cimpoi2016}, Quick Draw (Jongejan et al.,2016), Fungi \citep{Schroeder2018}, VGG Flower  \citep{Nilsback2008} and Traffic Signs \citep{Houben2013}. Despite having a small number of tasks, developing this benchmark opens the door to the use of multiple datasets for few-shot learning by leveraging the use of samples from different dataset to construct the prior knowledge from multiple data sources and allows researchers to evaluate more challenging generalization problem.

In the literature of genomics applications, most of the works are developing algorithms and techniques for only solving one particular clinical task at a time. To the best of our knowledge, there is no particular work which is doing general clinical feature prediction using gene expression data, although one of the works that has a similar initiative is "Multitask learning and benchmarking with clinical time-series data" by \citep{Harutyunyan2019}, they are only considering 4 different tasks from MIMIC-III database which is different from what we have done. This difference is coming from the number of defined tasks, the dataset that the tasks are originated from as well as defining a few-shot setting for each task in our Meta-Dataset. 

\section{Task Definition}
\label{sec:task-definition}
\begin{figure}
  \centering
   \begin{tabular}{@{}c@{}}
    \includegraphics[width=1.0\textwidth]{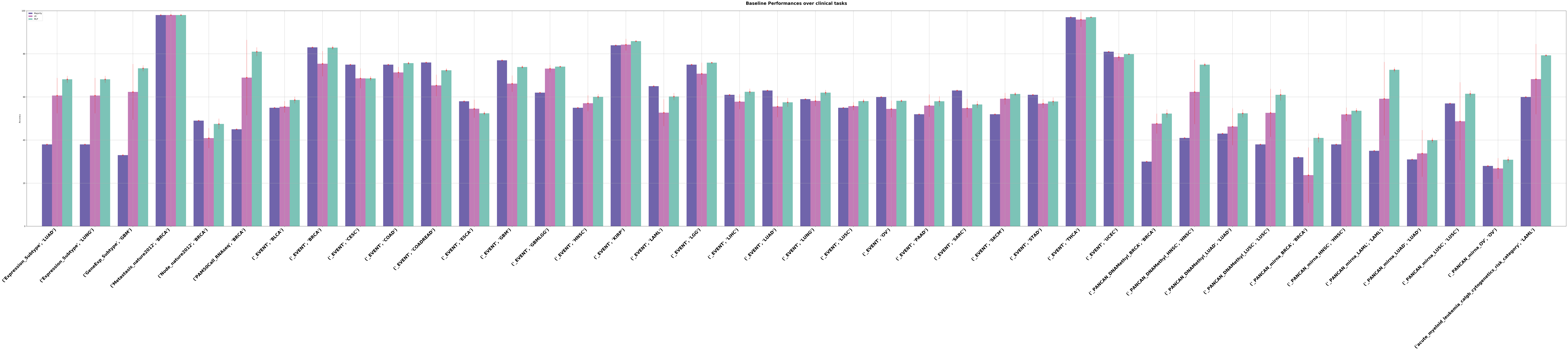}
  \end{tabular}
  \begin{tabular}{@{}c@{}}
    \includegraphics[width=1.0\textwidth]{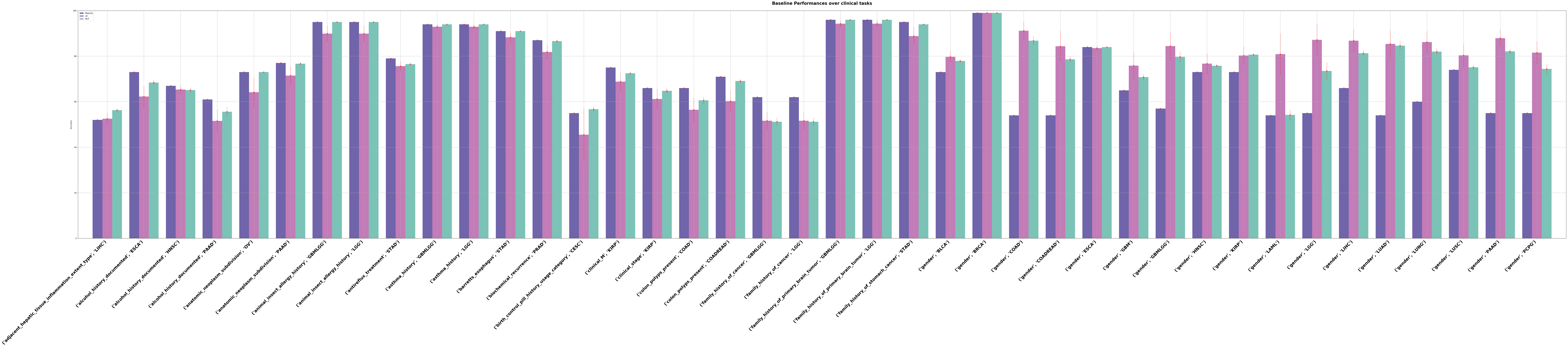}
  \end{tabular}
   \begin{tabular}{@{}c@{}}
    \includegraphics[width=1.0\textwidth]{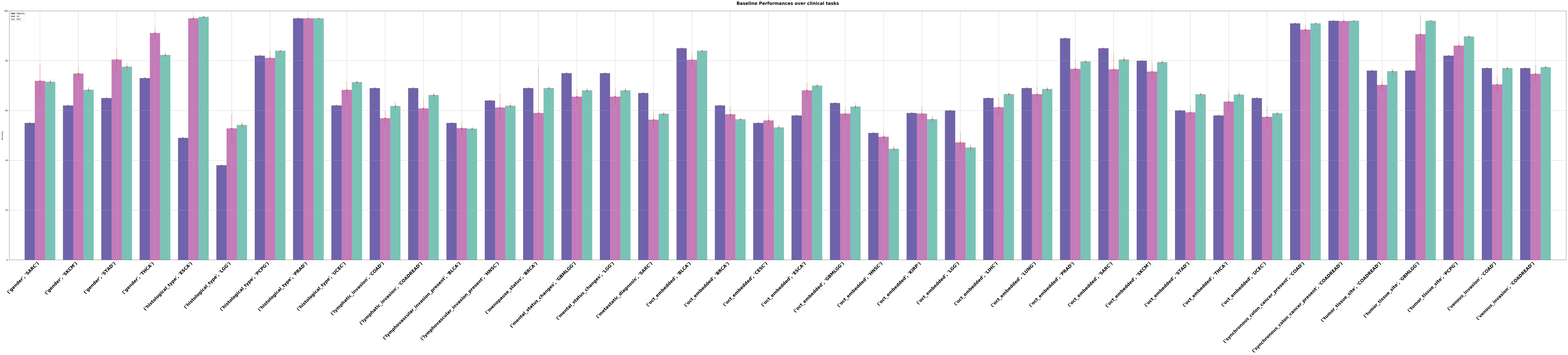}
  \end{tabular}
\caption{Model Performance over all datasets. Sub-figures show the performance of Majority classifier, Logistic Regression and MLP across all tasks.}
\label{fig:model-performance}
\end{figure}

The Cancer Genome Atlas (TCGA) program collected clinicopathologic annotation data along with multi-platform molecular profiles of more than 11,000 human tumors across 33 different cancer types \citep{Liu:TCGA:2018}. We use the gene expression profile (20530 genes) called IlluminaHiSeq that is obtained from RNA-Seq using some alignment and quantification algorithm. We consider all cancer studies available in TCGA to predict the phenotypes corresponding to them. To perform the prediction, we restrict ourselves to work only with samples for which TCGA has both RNA-Seq and a given phenotype.
There are 44 clinical variables (phenotypes) and 25 cancer studies. Tasks are combinations of a phenotype to predict and a cancer study. The defined tasks cover a range of clinical problems including modeling tumor tissue site, white cell count, histological type, family history of cancer, gender, pancan DNAMethyl which is the pan-cancer patterns of DNA methylation, pancan miRNA which is the pan-cancer data analysis of microRNA, animal allergy history, asthema history, mental status changes and a few others.
 You can find the full list of tasks in figure~\ref{fig:model-performance} and the Github Repo.  There are 174 tasks in total in which 121 of them have enough number of samples by considering the tasks that have at least 150 samples. We use gene expression data from tumor samples of a particular cancer study to predict a phenotype. For example, a task called ('$\_$EVENT', 'KIRP') means we predict a phenotype called \_EVENT which is an overall survival indicator (1=death, 0=censor) for TCGA Kidney Papillary Cell Carcinoma (KIRP) tumor samples. There are 174 tasks in TCGA Meta-Dataset, a full list of tasks besides the baselines' performance summary can be found in Figure ~\ref{fig:model-performance}.

\vspace*{-0.5em}
\section{Data-loaders for TCGA Meta-Dataset}
\label{sec:data-loaders-for-TCGA-Meta-Dataset}
We offer a meta-dataloader for TCGA that allows loading a batch of tasks that can be iterated over to generate datasets. We also provide the possibility of loading only one clinical task by specifying the phenotype and the cancer study. The file \texttt{TCGA.py} includes the detail of meta-dataloader. You can see an example below in which the variable 'datasets' is the collection of all tasks and 'task' is one of these datasets named (’PAM50Call$\_$RNAseq’, ’BRCA’). TCGA meta-dataloader is a part of Torchmeta Library \citep{deleu2019torchmeta} .\\[0.5em]
\begin{adjustbox}{center}
  \begin{tikzpicture}
    \node[font=\footnotesize\ttfamily, text width=\linewidth, inner sep=5pt, fill=CadetBlue!5, align=left] at (0, 0) {\textcolor{Gray}{\# Helper function, equivalent to Section~\ref{sec:data-loaders-for-TCGA-Meta-Dataset}}\\datasets = meta$\_$dataloader.TCGA.TCGAMeta(download=\textcolor{Green}{"True"}, min$\_$samples$\_$per$\_$class=\textcolor{Gray}{10})\\[0.5em]
    task = meta$\_$dataloader.TCGA.TCGATask(('PAM50Call$\_$RNAseq', 'BRCA'))\\
    [1em]\textcolor{OliveGreen}{print}(task.id)\textcolor{Gray}{ \# output: ('PAM50Call$\_$RNAseq', 'BRCA')}\\
    \textcolor{OliveGreen}{print}(task.\_samples.shape)\textcolor{Gray}{ \# output: (956, 20530)} \\
    \textcolor{OliveGreen}{print}(collections.Counter(task.$\_$labels))\textcolor{Gray}{ \# output: Counter({2: 434, 3: 194, 0: 142, 4: 119, 1: 67})}
    };
  \end{tikzpicture}
\end{adjustbox}
\vspace{-0.5cm}
\section{Experimental Results}
\label{sec:experimental-results}
We used three supervised models to evaluate the performance of each task using regression and neural network baselines. For the majority class predictions, we use a dummy classifier from Scikit-learn \citep{Pedregosa:2011:SML:1953048.2078195}, for logistic regression we use a 1-layer neural network (zero-hidden layer neural net with Softmax output) from PyTorch \citep{paszke2017automatic}. As the third baseline, after a hyperparameter search, we use a 3-layers network with 128 and 64 of hidden state size respectively, a learning rate equal to \(10^{-4}\), batch size of 32 and weight decay of 0.0 (no regularization effect). We use Adam \citep{Adam} for optimization in the neural network and LBFGS \citep{Liu:1989:LMB:3112655.3112866} for logistic regression. We run the network for 250 epochs for each task and over 10 different trials (by changing seed from 0 to 9). All tasks have 150 samples in total. We split 150 samples between train and test sets and valid set if needed. Train and test sets contain 50 and 100 samples respectively.
\vspace{-0.2cm}
\paragraph{Results}
MLP has the highest overall performance compared to Logistic Regression and Majority. The main limitation of logistic regression is that it cannot model the interactions between input variables. While the MLP is dealing better with clinical classification tasks, logistic regression has higher performance on gender tasks. In table~\ref{result-table}, you can see the average accuracy of each baseline over all tasks as well as the standard deviation. This result has been concluded over 10 repeats. 
\begin{table}
  \caption{Accuracy and standard deviation of 3 baselines over all tasks}
  \label{result-table}
  \centering
  \begin{tabular}{lll}
    \toprule
    \cmidrule(r){1-2}
    Model     & Accuracy     & Standard Deviation \\
    \midrule
    Majority &66.18  &16.87     \\
    Logistic Regression     &68.78 &15.91    \\
    Neural Network (MLP)     & 71.19  &14.62  \\
    \bottomrule
  \end{tabular}
\end{table}

\section{Conclusion and Future works}
\label{sec:conclusion-and-future-works}
We proposed a Meta-Dataset of tasks where each task is combinations of a clinical variable (phenotype) to predict and a cancer study. This dataset can be used as a testbed for developing algorithms for genomics applications which can leverage a few-shot learning regime. We evaluated the performance
of clinical tasks using supervised models and compared their accuracy to provide a baseline of what current techniques are able to achieve on the defined tasks.
For future work, we can apply meta-learning methods to construct the shared structural information within interrelated clinical tasks by using gene-expression data. As there is a lack of number of samples in the few-shot learning regime, by using meta-learning methods we leverage the samples from a collection of tasks and construct a prior knowledge for a general prediction. Also, studying the way that we can transfer these features from one clinical task to another could be an interesting direction.


\bibliographystyle{apalike}
\bibliography{references}

\end{document}